\pdfoutput=1

\documentclass[11pt]{article}

\usepackage[]{emnlp2021}

\usepackage{times}
\usepackage{latexsym}

\usepackage[T1]{fontenc}

\usepackage[utf8]{inputenc}

\usepackage{microtype}

\usepackage{bm}
\usepackage{graphicx}
\usepackage{amsmath}
\usepackage{multirow}
\usepackage{cleveref}
\usepackage{xspace}
\usepackage{xcolor}
\usepackage{enumitem}
\usepackage{placeins}

\newcommand{\caqa}{{CAQA}\xspace}

\title{Contrastive Domain Adaptation for Question Answering using Limited Text Corpora}

\author{Zhenrui Yue \\
  Technical University of Munich \\
  \texttt{zhenrui.yue@tum.de} \\\And
  Bernhard Kratzwald \\
  ETH AI Center \\
  \texttt{bkratzwald@ethz.ch} \\\And
  Stefan Feuerriegel \\
  ETH Zurich \\
  \texttt{sfeuerriegel@ethz.ch} \\}

\begin{document}
\maketitle
\begin{abstract}
Question generation has recently shown impressive results in customizing question answering (QA) systems to new domains. These approaches circumvent the need for manually annotated training data from the new domain and, instead, generate synthetic question-answer pairs that are used for training. However, existing methods for question generation rely on large amounts of synthetically generated datasets and costly computational resources, which render these techniques widely inaccessible when the text corpora is of limited size. This is problematic as many niche domains rely on small text corpora, which naturally restricts the amount of synthetic data that can be generated. In this paper, we propose a novel framework for domain adaptation called contrastive domain adaptation for QA (\caqa). Specifically, \caqa combines techniques from question generation and domain-invariant learning to answer out-of-domain questions in settings with limited text corpora. Here, we train a QA system on both source data and generated data from the target domain with a contrastive adaptation loss that is incorporated in the training objective. By combining techniques from question generation and domain-invariant learning, our model achieved considerable improvements compared to state-of-the-art baselines.
\end{abstract}

\section{Introduction}
\label{sec:introduction}

Question answering~(QA) systems generate answers to questions over text. Formally, such systems are nowadays trained end-to-end to predict answers conditional on an input question and a context paragraph~\cite[e.g.,][]{seo2016bidirectional, chen-etal-2017-reading, devlin-etal-2019-bert}. Therein, every QA sample is a 3-tuple consisting of a question, a context, and an answer. In this paper, we consider the subproblem of extractive QA, where the task is to extract answer spans from an unstructured context information for a given question as input. In extractive QA, both question and context are represented by running text, while the answer is defined by a start position and an end position in the context. 

An existing challenge for extractive QA systems is the distributional change between training data (source domain) and test data (target domain). If there is such a distribution change, the performance on test data is likely to be impaired. In practice, this issue occurs due to the fact that users, for instance, formulate text in highly diverse language or use QA for previously unseen domains~\cite{hazen2019towards, miller2020effect}. As a result, out-of-domain (OOD) samples occur that diverge from the training corpora of QA systems (i.e., which can be traced back to the invariance of the training data) and, upon deployment, lead to a drastic drop in the accuracy of QA systems. 

\begin{figure}[t]
    \centering
    \includegraphics[width=1.0\linewidth]{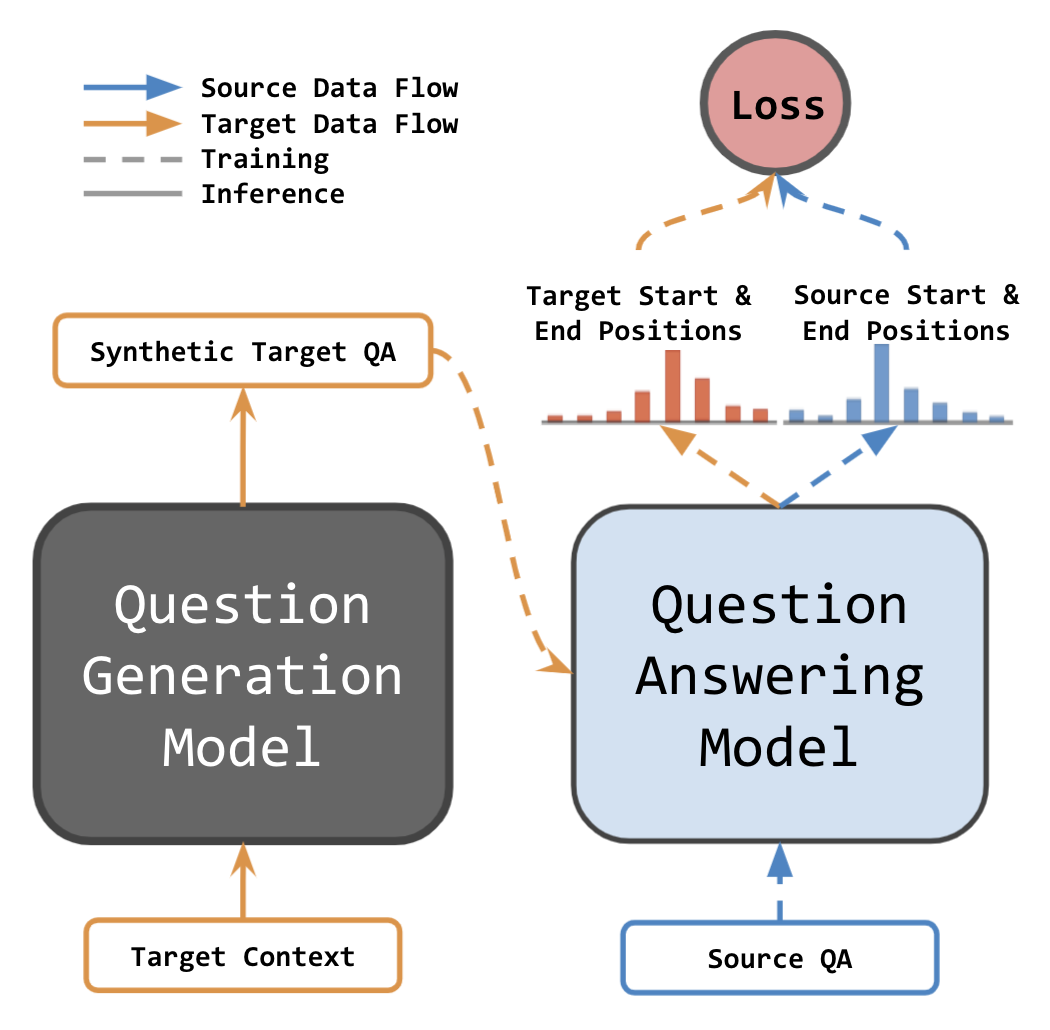}
    \caption{Overview of a common framework for QA domain adaptation. A question generation model is used to generate synthetic target data, which can be used for training the QA model with source data. The resulting QA model can answer target questions upon deployment.} 
    \label{fig:baseline}
\end{figure}

One solution to the above-mentioned challenge of a domain shift is to generate synthetic data from the corpora of the target domain using models for question generations and then use the synthetic data during training \cite[e.g.,][]{lee-etal-2020-generating, shakeri-etal-2020-end}. For this purpose, generative models have been adopted to produce synthetic data as surrogates from target domain, so that the QA system can be trained with both data from the source domain and synthetic data, which helps to achieve better results on the out-of-domain data distribution~\cite{puri-etal-2020-training, lee-etal-2020-generating, shakeri-etal-2020-end}, see \Cref{fig:baseline} for an overview of such approach. Nevertheless, large quantities of synthetic data require intensive computational resources. Moreover, many niche domains rely upon limited text corpora. Their limited size puts barriers to the amount of synthetic data that can be generated and, as well shall see later, render the aforementioned approach for limited text corpora largely ineffective.

In computer vision, some works draw upon another approach for domain adaptation, namely discrepancy reduction of representations~\cite{long2013transfer, tzeng2014deep, long2015learning, long2017deep, kang2019contrastive}. Here, an adaptation loss or adversarial training approaches are often designed to learn domain-invariant features, so that the model can transfer learnt knowledge from the source domain to the target domain. However, the aforementioned approach for domain adaptation was designed for computer vision tasks, and, to the best of our knowledge, has not yet been tailored for QA.

In this paper, we develop a framework for answering out-of-domain questions in QA settings with limited text corpora. We refer to our proposed framework as  \textbf{contrastive domain adaptation for question answering} (\caqa). \caqa combines question generation and contrastive domain adaptation to learn domain-invariant features, so that it can capture both domains and thus transfer knowledge to the target distribution. This is in contrast to existing question generation where synthetic data is solely used for joint training with the source data but without explicitly accounting for domain shifts, thus explaining why \caqa improves the performance in answering out-of-domain questions. For this, we propose a novel contrastive adaptation loss that is tailored to QA. The contrastive adaptation loss uses maximum mean discrepancy (MMD) to measure the discrepancy in the representation between source and target features, which is reduced while it simultaneously separates answer tokens for answer extraction.\footnote{The code from our \caqa framework is publicly available via \url{https://github.com/Yueeeeeeee/CAQA}}

The main \textbf{contributions} of our work are: 
\begin{enumerate}[leftmargin=15pt,itemsep=3px,nolistsep]
  \item We propose a novel framework for domain adaptation in QA called \caqa. To the best of our knowledge, this is the first use of contrastive approaches for learning domain-invariant features in QA systems. 
  \item Our \caqa framework is particularly effective for limited text corpora. In such settings, we show that \caqa can transfer knowledge to target domain without additional training cost. 
  \item We demonstrate that \caqa can effectively answer out-of-domain questions. \caqa outperforms the current state-of-the-art baselines for domain adaptation by a significant margin.
\end{enumerate}

\section{Related Work}
\label{sec:related_work}
The performance of extractive question answering systems \cite[e.g.,][]{chen2017reading, kratzwald-etal-2019-rankqa, zhang2020retrospective} is known to deteriorate when the training data (source domain) differs from the data used during testing (target domain) \cite{talmor-berant-2019-multiqa}. Approaches to adapt QA systems to a certain domain can be divided in (1)~supervised approaches, where one has access to labeled data from the target domain \cite[i.e., transfer learning;][]{kratzwald.2019a}, or (2)~unsupervised approaches, where no labeled information is accessible. The latter is our focus. Unsupervised approaches are primarily based on question generation techniques where one generates synthetic training data for the target domain. 

\textbf{Question generation (QG)}: Question generation~\cite{rus-etal-2010-first} is the task of generating synthetic QA pairs from raw text data. Several approaches have been developed to generate synthetic questions in QA. \citet{du-etal-2017-learning} propose an end-to-end seq2seq encoder-decoder for the generation. Recently, question generation and answer generation are observed as dual tasks and combined in various ways. \citet{tang2017question} train both simultaneously; \citet{golub-etal-2017-two} split the process in two consecutive stages; and \citet{tang-etal-2018-learning} use policy gradient to improve between-task learning.

Question generation is a common technique for domain adaptation in QA. Here, the generated questions are used to fine-tune QA systems to the new target domain \citep{dhingra-etal-2018-simple}. Oftentimes, only a subset of generated questions is selected to increase the quality of the generated data. Common approaches are based on curriculum learning \cite{sachan-xing-2018-self}; roundtrip consistency, where samples are selected when the predicted answers match the generated answer~\cite{alberti-etal-2019-synthetic}; iterative refinement~\cite{li-etal-2020-harvesting}; and conditional priors~\cite{lee-etal-2020-generating}.

\begin{figure*}[t]
    \centering
    \includegraphics[width=1.0\linewidth]{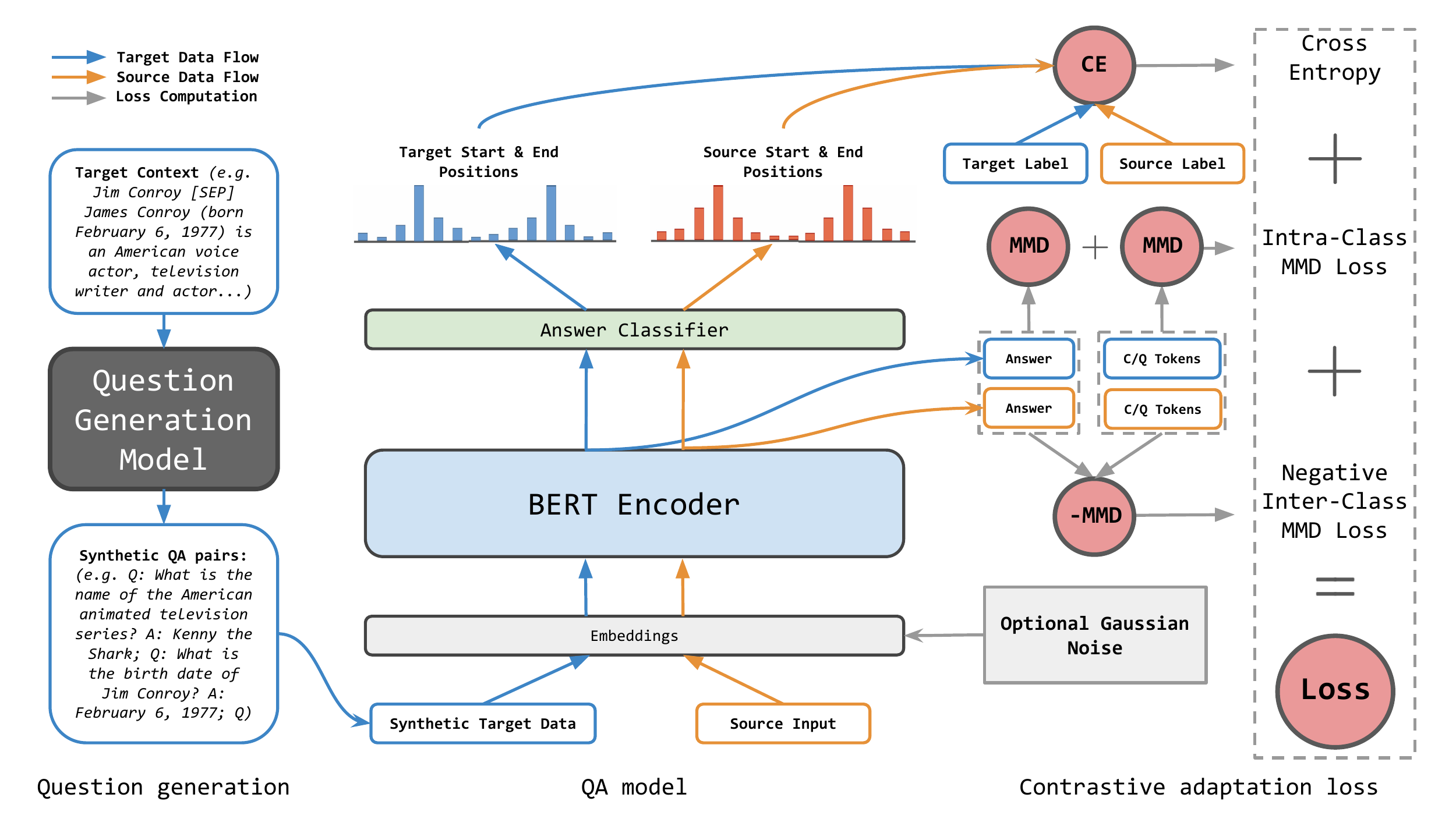}
    \caption{Overview of the proposed \caqa framework. A question generation model is used to generate synthetic data, which are then used for training the QA model using the contrastive adaptation loss. The resulting QA model is thus designed to handle QA data from the target domain upon deployment.} 
    \label{fig:intro}
\end{figure*}

\textbf{Unsupervised domain adaptation}: A large body of work on unsupervised domain adaptation has been done in the area of computer vision, where the representation discrepancy between a labeled source dataset and an unlabeled target dataset is reduced \cite[e.g.,][]{tzeng2014deep, saito2018maximum, long2015learning}. Recent approaches are often based on adversarial learning, where one minimizes the distance between feature distributions in both both the source and target domain, while simultaneously minimizing the error in the labeled source domain \cite[e.g.,][]{long2017deep,tzeng2017adversarial}. Moreover, adversarial training is also applied to train generalized QA systems across domains to improve performance on the data distribution of the target domain~\cite{lee-etal-2019-domain}.

Unlike adversarial approaches, contrastive methods \cite[i.e.,][]{hadsell2006dimensionality} utilize a special loss that reduces the discrepancy of samples from the same class (`pulled together') and that increases the distances for samples from different classes (`pushed apart'). This is achieved by using either pair-wise distance metrics \cite{hadsell2006dimensionality}, or a triplet loss and clustering techniques \cite{schroff2015facenet, cheng2016person}. Recently, a contrastive adaptation network (CAN) has been shown to achieve state-of-the-art performance by using maximum mean discrepancy to build an objective function that maximizes inter-class distances and minimizes intra-class distances with the help of pseudo-labeling and iterative refinement \cite{kang2019contrastive}. Yet, it is hitherto unclear how this technique can be used to improve domain adaptation in QA.

\section{The \caqa Framework: Contrastive Domain Adaptation for QA}
\label{sec:methodology}

\subsection{Setup}
\textbf{Input}: Our framework is based on QA data under a distributional change. Thus let $\bm{\mathcal{D}}_{s}$ denote the source domain and $\bm{\mathcal{D}}_{t}$ the target domain. We expected a distributional change, that is, both domains are different (i.e., $\bm{\mathcal{D}}_{s} \neq \bm{\mathcal{D}}_{t}$). Formally, the input is given by:
\begin{itemize}[leftmargin=10pt]
    \item \emph{Training data from source domain}: We are given labeled data from the source domain $\bm{X}_{s}$. Each sample $\bm{x}_{s}^{(i)} \in \bm{X}_{s}$ from the source domain $\bm{\mathcal{D}}_{s}$ comprises of a 3-tuple with a question $\bm{x}_{s,q}^{(i)}$, a context $\bm{x}_{s,c}^{(i)}$, and an answer $\bm{x}_{s,a}^{(i)}$.
    \item \emph{Target contexts}: We have access to target domain data. Yet, of note, the data is unlabeled. That is, we have only access to the contexts. We further assume that the amount of target contexts is limited. Let $\bm{X}_{t}^{'}$ denote the unlabeled target data, where each sample $\bm{x}_{t}^{(i)} \in \bm{X}_{t}^{'}$ from the target domain $\bm{\mathcal{D}}_{t}$ consists of only a context $\bm{x}_{t,c}^{(i)}$.
\end{itemize}
\textbf{Objective:} Upon deployment, we aim at maximizing the performance of the QA system when answering questions from the target domain $\bm{\mathcal{D}}_{t}$, that is, minimizing the cross-entropy loss of the QA system $\bm{f}$ for $\bm{X}_{t}$ from the target domain $\bm{\mathcal{D}}_{t}$, i.e.,
\begin{equation}
    \bm{f}^{*} = \arg \min_{\substack{\bm{f}}} \sum_{i=1}^{|\bm{X}_{t}|} \mathcal{L}_{\mathrm{ce}}(\bm{f}(\bm{x}_{t,c}^{(i)}, \bm{x}_{t,q}^{(i)}), \bm{x}_{t,a}^{(i)}). \label{eq:objective}
\end{equation}
However, actual question-answer pairs from the target domain are unknown until deployment. Furthermore, we expect that the available contexts are limited in size, which we refer to as limited text corpora. For instance, our experiments later involve only 5 QA pairs per context and, overall, 10k paragraphs as context.

\textbf{Overview:} The proposed \caqa framework has three main components (see \Cref{fig:intro}): (\textsc{1})~a \textbf{question generation model}, (\textsc{2})~a \textbf{QA model}, and (\textsc{3})~a \textbf{contrastive adaptation loss} for domain adaptation, as described in the following. We refer to the question generation model via $\bm{f}_{\mathrm{gen}}$ and to the QA model via $\bm{f}$. The question generation model $\bm{f}_{\mathrm{gen}}$ is used for generate synthetic QA data $\bm{X}_{t} = \bm{f}_{\mathrm{gen}}(\bm{X}_{t}^{'})$. This yields additional QA pairs consisting of $\bm{x}_{t,q}^{(i)}$ and $\bm{x}_{t,a}^{(i)}$ for $\bm{x}_{t,c}^{(i)} \in \bm{X}_{t}^{'}$. Then, we use both source data $\bm{X}_{s}$ and synthetic data $\bm{X}_{t}$ to train the QA model via our proposed contrastive adaptation loss. The idea behind it is to help transfer knowledge to the target domain via discrepancy reduction and answer separation. 

\subsection{Question Generation}
\label{section:qg-model}

The question generation (QG) model QAGen-T5 is designed as follows. The QG model takes a context as input and then involves two steps: (i)~it first generates a question $\bm{x}_{q}$ based on context $\bm{x}_{c}$ in the target domain, and then (ii)~a corresponding answer $\bm{x}_{a}$ conditioned on given $\bm{x}_{c}$ and $\bm{x}_{q}$. Using a two-step generation of questions and answers to build synthetic data is consistent with earlier literature on QG~\cite[e.g.,][]{lee-etal-2020-generating, shakeri-etal-2020-end} and thus facilitates larger capacity while facilitating comparability. The maximum number $k$ of synthetic QA data is determined later.  

In our QG model, we utilize a text-to-text transfer transformer (T5) encoder-decoder transformer \cite{raffel2019exploring}. This transformer is able of performing multiple downstream tasks due to its the multi-task pretraining approach. This is beneficial in our case as we later use T5 transformers for conditional generation of two different outputs $\bm{x}_{q}$ and $\bm{x}_{a}$, respectively. Specifically, we use two T5 transformers for generating end-to-end (i)~the question and (ii)~the answer. We later refer to the combined variant for QG as `QAGen-T5'.

Our QAGen-T5 is fed with the following input/output. The \textbf{input} to generate questions is only a context paragraph, and, therefore, we prepend the token \texttt{generate question:} in the beginning (which is then followed by the context paragraph). For answer generation, \textbf{input} using both a question and a context is specified via tokens \texttt{question:} and \texttt{context:}. The \textbf{output} varies across (i)~question and (ii)~answer. For (i), the output $\bm{x}_{q}$ are questions divided by the \texttt{[SEP]} token (e.g., input: `\emph{generate question: python is a programming language...}' output: `\emph{when was python released?}'). For (ii), the output $\bm{x}_{a}$ is an answer, for which we specify question and context information in the input by inserting tokens \texttt{question:} and \texttt{context:} (e.g., the input becomes `\emph{question: when was python released? context: python is a programming language...}'). The output is the decoded answer. 

QAGen-T5 is trained as follows. For (i) and (ii), we separately minimize the negative log likelihood of output sequences via
\begin{align}
    \mathcal{L}_{\mathrm{qg}}(\bm{X}) &= \sum_{i=1}^{|\bm{X}|} -\log p_{\theta_{\mathrm{qg}}} (\bm{x}_{q}^{(i)} \mid \bm{x}_{c}^{(i)}), \label{eq:qg1} \\
    \mathcal{L}_{\mathrm{ag}}(\bm{X}) &= \sum_{i=1}^{|\bm{X}|} -\log p_{\theta_{\mathrm{ag}}} (\bm{x}_{a}^{(i)} \mid \bm{x}_{c}^{(i)}, \bm{x}_{q}^{(i)}), \label{eq:qg2}
\end{align}
where $\bm{x}_{q}^{(i)}$, $\bm{x}_{a}^{(i)}$, and $\bm{x}_{c}^{(i)}$ refer to question, answer, and context in the $i$-th sample of $\bm{X}$. Fine-tuning is done as follows. Both T5 models inside QAGen-T5 are fine-tuned on SQuAD separately. For selecting QA pairs, we draw upon LM-filtering~\cite{shakeri-etal-2020-end} to select the best $k$ QA pairs per context (e.g., $k=5$ is selected later in our experiments). We compute the LM scores for the answer by multiplying the scores of each token over the output length. This ensures that only synthetic QA samples are generated where the combination of both question and answer has a high likelihood.   

\subsection{QA Model}

Our QA model is set to BERT-QA~\cite{devlin-etal-2019-bert}. BERT-QA consists of two components: the BERT-encoder and an answer classifier. The BERT-encoder extracts features from input tokens, while the answer classifier outputs two probability distributions for start and end positions to form answer spans based on the token features extracted by BERT-encoder. In our paper, the BERT-encoder is identical to the original BERT model and has an embedding component as well as transformer blocks~\cite{devlin-etal-2019-bert}.

BERT-QA is trained using a cross-entropy loss $\mathcal{L}_{\mathrm{ce}}$ to predict correct answer spans, yet additionally using our contrastive adaptation loss as described in the following.

\subsection{Contrastive Adaptation Loss}
\begin{figure}[t]
    \centering
    \includegraphics[width=1.0\linewidth]{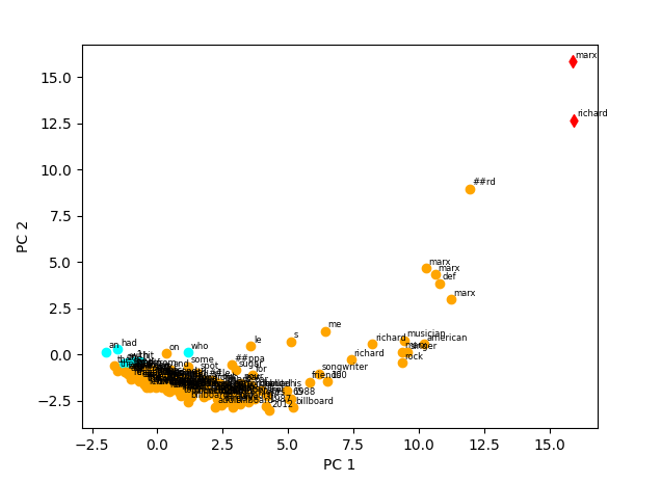}
    \caption{The 2D PCA visualization of BERT-encoder output on a SQuAD example. Answer tokens are in red and question tokens in cyan (all other tokens in orange).} 
    \label{fig:pca-example}
\end{figure}

We now introduce our contrastive adaptation loss, which we use for training the QA model. The idea in our proposed contrastive adaptation loss is two-fold: (i)~We decrease the discrepancy among answer tokens and among other tokens, respectively (`intra-class'). This should thus encourage the model to learn domain-invariant features that are characteristic for both the source domain and the target domain. (ii)~We enlarge the answer--context and answer--question discrepancy in feature representations (`inter-class').

Our approach is somewhat analogous yet different to contrastive domain adaptation in computer vision, where also the intra-class discrepancy is reduced, while the inter-class discrepancy is enlarged \cite{kang2019contrastive}. In computer vision, the labels are clearly defined (e.g., object class), such labels are not available in QA. A natural way would be to see each pair of start/end location of an answer span as a separate class. Yet the corresponding space would be extremely large and would not represent specific semantic information. Instead, we build upon a different notion of classes: we treat all answer tokens as one class and the combined set of question and context tokens as a separate class. When we then reduce intra-class discrepancy and enlarge inter-class discrepancy, knowledge is transferred from source to target domain.

We focus on the discrepancy between answer and the other tokens since a trained QA model can well separate answer tokens in the source domain; see \Cref{fig:pca-example}. The plot shows a principle component analysis (PCA) visualizing the BERT-encoder output of a SQuAD example~\cite{van2019does}. Answer tokens are well separated from all other tokens in this case, nevertheless, the same QA model can fail to perform answer separation in an unseen domain; see examples in \Cref{sec:pca-examples}. Therefore, we apply contrastive adaptation on the token level and define classes by token types. Ideally, this reduces feature discrepancy between domains and, by separating answer tokens. Both should help improving the performance on out-of-domain data.

\textbf{Discrepancy:} In our contrastive adaptation loss, we measure the discrepancy among token classes using maximum mean discrepancy (MMD). MMD measures the distance between two data distributions based on the samples drawn from them~\cite{gretton2012kernel}. Empirically, we compute the distance $D$ between tokens $X$ and $Y$ represented by their mean embeddings in reproducing kernel Hilbert space $\mathcal{H}$, i.e.,
\begin{equation}
  D = \sup_{f \in \mathcal{H}} \big( \frac{1}{|\bm{X}|} \sum_{i=1}^{|\bm{X}|}f(\bm{x}_{i}) - \frac{1}{|\bm{Y}|} \sum_{i=1}^{|\bm{Y}|}f(\bm{y}_{i}) \big). \label{eq:mmd}
\end{equation}
MMD can be simplified by choosing a unit ball in $\mathcal{H}$, such that $D(\bm{X}, \bm{Y})^{2} = \| \bm{\mu}_{x} - \bm{\mu}_{y} \|_{\mathcal{H}}^{2}$, where $\bm{\mu}_{x}$ and $\bm{\mu}_{y}$ represent the sample mean embeddings. Similar to~\cite{long2015learning}, we adopt Gaussian kernel with multiple bandwidths to estimate distances two samples, i.e., $k(\bm{x}_{i}, \bm{x}_{j}) = \mathrm{exp}(- \frac{\| \bm{x}_{i} - \bm{x}_{j} \|^{2}}{\gamma})$.

\textbf{Contrastive adaptation loss:} We define the contrastive adaptation loss of a mixed batch $\bm{X}$ with samples from both the source domain and the target domain as
\begin{equation}
  \begin{aligned}
    \mathcal{L}_{\mathrm{con}}(\bm{X}) &= \frac{1}{|\bm{X}|^{2}} \sum_{i=1}^{|\bm{X}|} \sum_{j=1}^{|\bm{X}|} k(\phi(\bm{x}_\mathrm{a}^{(i)}), \phi(\bm{x}_\mathrm{a}^{(j)})) \\
    &+ \frac{1}{|\bm{X}|^{2}} \sum_{i=1}^{|\bm{X}|} \sum_{j=1}^{|\bm{X}|} k(\phi(\bm{x}_\mathrm{cq}^{(i)}), \phi(\bm{x}_\mathrm{cq}^{(j)})) \\
    &- \frac{1}{|\bm{X}|^{2}} \sum_{i=1}^{|\bm{X}|} \sum_{j=1}^{|\bm{X}|} k(\phi(\bm{x}_\mathrm{a}^{(i)}), \phi(\bm{x}_\mathrm{cq}^{(j)})), \label{eq:contrastive_loss}
  \end{aligned}
\end{equation}
where $\bm{x}_\mathrm{a}$ is the mean vector of answer tokens, while $\bm{x}_\mathrm{cq}$ is the mean vector of the context/question tokens. Further, $\phi$ is a feature extractor (i.e., the BERT-encoder). \Cref{eq:contrastive_loss} fulfills our objectives (i) and (ii) from above. The first two terms estimate the mean distance among all answers tokens and the other tokens, respectively, This should thus fulfill objective (i): to minimize the intra-class discrepancy. The last term maximizes the distance between answer and rest tokens (i.e., by taking the negative distance) and enables an easier answer extraction. This should thus fulfill objective (ii): to maximize the inter-class discrepancy.

\textbf{Overall objective}: We now combine both the cross-entropy loss from BERT-QA and the above contrastive adaptation loss into a single optimization objective for the QA model:
\begin{equation}
  \mathcal{L}_{\mathrm{qa}}(\bm{X}) = \mathcal{L}_{\mathrm{ce}}(\bm{X}) + \beta \mathcal{L}_{\mathrm{con}}(\bm{X}),
\end{equation}
where $\mathcal{L}_{\mathrm{ce}}$ is the cross-entropy loss for the training QA model to predict correct answer spans. Here, $\beta$ is hyperparameter that we choose empirically.

In our experiments, we sample mixed mini-batches and compute the overall loss to update the QA model. We encourage correct answer extraction by maximizing the representation distance between answer tokens and the other tokens. Additionally, we apply Gaussian noise at different scales $\sigma$ on the token embeddings to learn a smooth and generalized feature space~\cite{cai2019learning}.

\section{Experiment Setup}
\label{sec:experiments}

\begin{table*}[t]
  \centering
  \begin{tabular}{llcccc}
  \hline
  \multirow{2}{*}{\textbf{Model}} & \multirow{2}{*}{\textbf{Training data}} & \textbf{TriviaQA}    & \textbf{HotpotQA}    & \textbf{NaturalQ.}  & \textbf{SearchQA}    \\
                                  &                                         & EM / F1              & EM / F1              & EM / F1              & EM / F1              \\ \hline
  \multicolumn{6}{l}{\qquad\qquad(\textsc{I}) Performance on target datasets w/o domain adaptation}                                                                                              \\ \hline
  BERT-QA                         & SQuAD                                   & 50.84/60.48          & 43.57/61.09          & 45.14/59.35          & 19.66/27.90          \\ \hline
  \multicolumn{6}{l}{\qquad\qquad(\textsc{II}) Performance on target datasets w/ supervised training}                                                                                             \\ \hline
  BERT-QA                         & 10k Target                              & 55.86/62.14          & 49.30/66.23          & 55.26/68.19          & 62.48/69.39          \\
  BERT-QA                         & All Target                              & 65.28/70.80          & 57.69/74.78          & 67.25/79.03          & 72.46/78.76          \\ \hline
  \multicolumn{6}{l}{\qquad\qquad(\textsc{III}) Performance on target datasets w/ question generation}                                                                                             \\ \hline
  Info-HCVAE                      & S + 10k Syn                             & 45.66/55.28          & 39.47/55.60          & 37.12/51.17          & 17.41/24.24          \\
  AQGen                           & S + 10k Syn                             & 51.41/60.60          & 44.79/59.99          & 40.78/54.02          & 34.03/42.08          \\
  QAGen                           & S + 10k Syn                             & 50.52/59.79          & 45.67/60.88          & 44.13/57.84          & 29.59/36.63          \\
  QAGen-T5 (proposed)             & S + 10k Syn                             & 54.32/62.74          & \textbf{46.50/62.03} & 46.48/60.65          & 32.54/39.44          \\
  \caqa (proposed)                & S + 10k Syn                             & \textbf{55.17/63.23} & 46.37/61.57          & \textbf{48.55/62.60} & \textbf{36.05/42.94} \\ \hline
  \end{tabular}
  \caption[Main results]{Main results comparing question-answering performance on out-of-domain data.}
  \label{tab:main-results}
\end{table*}

\subsection{Datasets}
In our experiments, we use SQuAD v1.1 as our \textbf{source domain} dataset~\cite{rajpurkar-etal-2016-squad}. For the target domain we adopt four other datasets from MRQA~\cite{fisch-etal-2019-mrqa}. This allows us to evaluate the performance in answering out-of-domain questions. For \textbf{target domain} datasets, only context paragraphs are accessible for question generation. In this paper, we use TriviaQA \cite{joshi-etal-2017-triviaqa}, HotpotQA \cite{yang-etal-2018-hotpotqa},  NaturalQuestions~\cite{kwiatkowski-etal-2019-natural}, and {SearchQA}~\cite{dunn2017searchqa}. For more details, see \Cref{sec:datasets}.

\subsection{Baselines}
We draw upon the following state-of-the-art baselines for question generation: Info-HCVAE~\cite{lee-etal-2020-generating}, AQGen~\cite{shakeri-etal-2020-end}, and QAGen~\cite{shakeri-etal-2020-end}. These are used to generate synthetic QA data in order to train BERT-QA as the underlying QA model (i.e., the QA model is the same as in \caqa, the only difference is in how synthetic QA data is generated and how the QA model is trained). For more details, see \Cref{sec:baselines}. 

Additionally, we train BERT-QA~\cite{devlin-etal-2019-bert} on SQuAD and target datasets, respectively. This is to evaluate its base performance with zero knowledge of the target domain (`lower bound') and supervised training performance (`upper bound') on target datasets. We further report QAGen-T5 as an ablation study reflecting \caqa without the contrastive adaptation loss.  

\subsection{Training and Evaluation}
\textbf{Training:}

We perform experiments based on limited text corpora: we only allow 5 QA pairs per context and 10k context paragraphs in total to be generated as the surrogate dataset. As such, no intensive computational resources are required for QA domain adaptation. First, we randomly select 10k context paragraphs and generate QA pairs with all mentioned generative models as synthetic data (abbreviated as `10k Syn'). Then, QA pairs are filtered using roundtrip consistency (baseline models) or LM-filtering (QAGen-T5), such that max. 5 QA pairs are preserved for each context. The final training data is then given by the combination of the generated target QA pairs and the SQuAD training set (`S + 10k Syn'). Based on this, we train BERT-QA on it and evaluate the model on target dev sets.

\textbf{Evaluation:} For evaluation, we adopt two metrics: exact match (EM) and F1 score (F1). We compute these metrics on target dev sets to evaluate the out-of-domain performance of the trained QA systems. For details, see \Cref{sec:implementation}.

\section{Results}

\subsection{Performance on Target Questions}
\Cref{tab:main-results} reports the main results. The table includes several baselines: (\textsc{I})~BERT-QA using only SQuAD as na{\"i}ve baseline without domain adaptation (`lower bound'); (\textsc{II})~BERT-QA with supervised learning and thus access to the target data, which are otherwise not used (`upper bound'); and (\textsc{III})~several state-of-the-art baselines for question generation. 

We make the following observations: (1)~The na{\"i}ve baseline is consistently outperformed by our proposed \caqa framework. Compared to the SQuAD baseline, \caqa leads to a performance improvement in EM of at least $2.80\%$ and can go as high as $16.39\%$, and an improvement in F1 of at least $0.48\%$ and up to $15.04\%$. (2)~The na{\"i}ve baseline provides a challenge for several question generation baselines from the literature, which are often inferior in our setting with limited text corpora. (3)~The best-performing approach is our \caqa framework for three out of four datasets. For one dataset (HotpotQA), it is our QAGen-T5 variant without contrastive adaptation loss. However, the performance of \caqa is of similar magnitude and is clearly ranked second. (4)~By comparing \caqa and QAGen-T5, we yield an ablation study quantifying the gain that should be attributed to using a contrastive adaptation loss. Here, we find distinctive performance improvements due to our contrastive adaptation loss for three out of four datasets. (5)~Compared to the question generation baselines, our \caqa framework is superior. Compared to AQGen, the average improvements in EM and F1 are 3.78\% and 3.41\%, respectively, and, compared QAGen, the average improvements are 4.06\% and 3.80\%, respectively. (6)~In the case of TriviaQA and HotpotQA, the performance of \caqa is close to that of supervised training results using 10k paragraphs from the target datasets. We discuss reasons reasons for performance variations across datasets in \Cref{sec:discussion}. 

Altogether, the results suggest that the proposed combination of QAGen-T5 and contrastive adaptation loss is effective in improving the performance for out-of-domain data.

\subsection{Sensitivity Analysis for Text Corpora Size}
\label{section:corpora-size-analysis}

We now perform a sensitivity analysis studying how the performance varies across different text corpora sizes, that is, the number of context paragraphs generated. For this, we randomly select 10k, \ldots, 50k context paragraphs for training and then report two variants: (i)~the QG performance using QAGen-T5 with varying context numbers and (ii) our CAQA with 10k context paragraphs. Here, results are reported for TriviaQA and NaturalQuestions\footnote{A sensitivity analysis varying the number of QA pairs ($k$) is reported using HotpotQA and SearchQA in \Cref{sec:varying_qa_per_context}}. 

For QAGen-T5, we see a comparatively large improvement when increasing the size from 10k to 20k context paragraphs. A small performance improvement among QAGen-T5 can be obtained when choosing 50k context paragraphs. In contrast to that, \caqa is superior, even when using only 10k context paragraphs. Put simply, it does so much with much fewer samples and thus without additional costs due to extra computations. In sum, this demonstrates the effectiveness of \caqa for improving QA domain adaptation in settings with limited text corpora.

\begin{table}[ht]
  \footnotesize
  \centering
  \begin{tabular}{llcc}
  \hline
  \multirow{2}{*}{\textbf{Model}} & \multirow{2}{*}{\textbf{Training data}} & \textbf{TriviaQA}    & \textbf{NaturalQ.}    \\
                                          && EM / F1              & EM / F1              \\ \hline
  \multicolumn{4}{c}{Performance w/o domain adaptation}                                 \\ \hline
  BERT-QA & SQuAD                                   & 50.84/60.48          & 45.14/59.35          \\ \hline
  \multicolumn{4}{c}{Performance w/ question generation}                                \\ \hline
  QAGen-T5 & S + 10k Syn                             & 54.32/62.74          & 46.48/60.65          \\
  QAGen-T5 & S + 20k Syn                             & 53.73/62.45          & 47.98/61.90          \\
  QAGen-T5 & S + 30k Syn                             & 54.75/63.12          & 47.76/61.93          \\
  QAGen-T5 & S + 40k Syn                             & 54.82/63.09          & 48.47/62.55          \\
  QAGen-T5 & S + 50k Syn                             & 54.90/63.11          & 48.23/62.71          \\
  \caqa & S + 10k Syn                     & \textbf{55.17/63.23} & \textbf{48.55/62.60} \\ \hline
  \end{tabular}
   \caption[Sensitivity analysis for text corpora size]{Sensitivity analysis across different corpora sizes. Top: QAGen-T5 w/o contrastive adaptation loss; below: \caqa with such loss.}
  \label{tab:context-number-results}
\end{table}

\subsection{Comparison: Training Baselines with Contrastive Adaptation Loss}  

We perform a sensitivity analysis examining whether the baselines models (Info-HCVAE, AQGen, and QAGen) can be improved when training them using our contrastive domain adaptation. For this, we repeat the above experiments with 10k synthetic samples (i.e., S + 10k Syn). The only difference is that we use our contrastive adaptation loss. The results are in \Cref{tab:baseline-contrastive-results}. Here, a positive value means that the use of a contrastive adaptation loss results in a performance gain (since everything else is kept equal). Note that combining QAGen-T5 with our contrastive adaptation loss yields \caqa. Overall, we see that the performance of almost baselines can be improved due to our contrastive adaptation loss. 

\begin{table}[ht]
  \footnotesize
  \centering
  \begin{tabular}{lcc}
  \hline
  \multirow{2}{*}{\textbf{Model}} & \textbf{TriviaQA}    & \textbf{HotpotQA}    \\
                                  & EM / F1              & EM / F1              \\ \hline
  Info-HCVAE                      & -0.33/-0.45          & +0.03/+0.01          \\
  AQGen                           & +0.21/+0.26          & \textbf{+0.86/+0.88} \\
  QAGen                           & -0.09/-0.37          & +0.42/+0.37          \\
  QAGen-T5 (=\caqa)                 & \textbf{+0.85/+0.49} & +0.70/+0.69          \\ \hline
  \multirow{2}{*}{\textbf{Model}} & \textbf{NaturalQ.}    & \textbf{SearchQA}    \\
                                  & EM / F1              & EM / F1              \\ \hline
  Info-HCVAE                      & +0.66/+0.47          & +1.01/+1.04          \\
  AQGen                           & +1.90/+1.85          & -0.25/-0.28          \\
  QAGen                           & +1.23/+0.79          & \textbf{+4.89/+5.50} \\
  QAGen-T5 (=\caqa)                 & \textbf{+2.07/+1.95} & +3.51/+3.50          \\ \hline
  \end{tabular}
  \caption[Improvement due to contrastive adaptation loss]{Performance improvements (absolute) when training baselines with our proposed contrastive adaptation loss.}  
  \label{tab:baseline-contrastive-results}
\end{table}

\section{Discussion}
\label{sec:discussion}

We now discuss variations in the performance across models and datasets. For this, we also investigate synthetic data generated by \caqa manually (see \Cref{sec:qualitative-examples}).

\vspace{0.2cm}
\noindent
\emph{Why is the performance sometimes below the upper bound (i.e., supervised training)?}

\noindent
We see two explanations for the performance gap between supervised training and \caqa (as well as the other baselines). {(i)}~Despite domain adaptation, some of the generated synthetic data cannot perfectly match the characteristics of the target domain but still reveal differences. We found this behavior, e.g., for NaturalQuestions. Here, the average length of the synthetic answers are all below 3, as compared to 4.35 the training set of NaturalQuestions. This may lead to a performance gap at test time. {(ii)}~The generated QA pairs are comparatively homogeneous and lack the diversity of the target domain. To examine this, we manually inspected synthetic samples from \caqa (see \Cref{sec:qualitative-examples}). We found that the generated QA pairs cannot fully capture the diversity that is otherwise common in question formulation. For example, almost all questions in the synthetic data start with ‘\emph{What}’, ‘\emph{When}’, and ‘\emph{Who}’. In contrast, in NaturalQuestions, we find many questions that we perceived as more diverse or event more difficult. Examples are ‘\emph{The court aquitted Moninder Singh Pandher of what crime?}’ and ‘\emph{Why does queen elizabeth sign her name elizabeth r?}’. Such behavior is particularly exacerbated for NaturalQuestions, which was intentionally designed to introduce more variety in question formulation, and, hence, our contrastive domain adaptation approach might implicitly learn some of the characteristics (as compared to the state-of-the-art baselines). 

\vspace{0.2cm}
\noindent
\emph{Why does the performance improvements vary across datasets?}\\
\noindent
The different improvements with contrastive adaptation can be further attributed to the target domain itself. When source and target datasets are similar, a model trained on the source dataset would naturally have better performance on the target dataset, but the improvements with contrastive adaptation can be limited due to the small domain variation. In TriviaQA and HotpotQA, the context paragraph originates -- partially or completely -- from Wikipedia and answer lengths are similar. In contrast, NaturalQuestions have different text styles and sources including raw HTML like `<Table>`, SearchQA context involves web articles and user contents, their average answer lengths are different, amounting to 1.89 and 4.43 respectively. Additionally, supervised training results using 10k HotpotQA and TriviaQA yield moderate improvements (5.02\%, 5.73\%), compared to 10.12\% and 42.82\% in NaturalQuestions and SearchQA. This also suggests that the difference between the previous datasets and SQuAD is comparatively small. Similar trends can be found in \Cref{tab:baseline-contrastive-results}, where our contrastive adaptation on baseline models proves to be more effective in NaturalQuestions and SearchQA. Therefore, the discrepancy between source domain and target domain can be crucial for domain adaptation results according to our observations.

\vspace{0.2cm}
\noindent
\emph{How does our contrastive adaptation loss affect the discrepancy among answer tokens?}\\
\noindent
To further examine how the contrastive adaptation loss improves the discrepancy among answers, we draw upon methods in \cite{van2019does} and visualize the representations of the answer tokens using PCA (see \Cref{sec:pca-examples}). Based on it, we empirically make the following observations. {(i)}~In correct predictions, answer tokens are separated very well from questions and context tokens. {(ii)}~In incorrect predictions, the correct answer is either not separated from the other tokens, or wrong tokens are separated and predicted as answers. In the latter case, such behavior is termed as overconfidence in out-of-domain data \citep[cf.][]{kamath-etal-2020-selective}. In sum, contrastive adaptation helps in separating tokens that are likely to be answers, though sometimes incorrect tokens are identified as answers, thereby worsening the problem of overconfidence, which may explain the occasional decrease in performance.

\section{Conclusion}
\label{sec:conclusion}

This work contributes a novel framework for domain adaptation of QA systems in settings with limited text corpora. We develop \caqa in which we combine techniques from from question generation and domain-invariant learning to answer out-of-domain questions. Different from existing works in question answering, we achieve this by proposing a contrastive adaptation loss. Extensive experiments show that \caqa is superior to other state-of-the-art approaches by achieving a substantially better performance on out-of-domain data.


\bibliography{anthology, custom}
\bibliographystyle{acl_natbib}

\clearpage
\appendix
\section{Implementation Details}
\subsection{Datasets}
\label{sec:datasets}
In our experiments we used the following datasets, for target datasets, we adopt modified versions from MRQA~\cite{fisch-etal-2019-mrqa}.
\begin{enumerate}[itemsep=3px,nolistsep,leftmargin=15pt]
\item \emph{SQuAD}~\cite{rajpurkar-etal-2016-squad} is a crowdsourced dataset with context passages from Wikipedia and human-labeled question-answer pairs. The adopted SQuAD v1.1 training set has 18,885 context paragraphs and 86,588 QA pairs.
\item \emph{TriviaQA}~\cite{joshi-etal-2017-triviaqa} is a large-scale QA dataset that includes QA pairs and supporting facts for supervised training.
\item \emph{HotpotQA}~\cite{yang-etal-2018-hotpotqa} provides multi-hop questions with human annotations (distractor paragraphs are excluded).
\item \emph{NaturalQuestions}~\cite{kwiatkowski-etal-2019-natural} contains actual questions from users issued to the Google search engine. In the MRQA version, short answers are adopted, while long answers are used as context paragraphs.
\item \emph{SearchQA}~\cite{dunn2017searchqa} was constructed through a pipeline that starts from existing QA pairs and search for context information based on crawled online search results.
\end{enumerate}

\subsection{Baselines}
\label{sec:baselines}
\begin{enumerate}[leftmargin=15pt,itemsep=3px,nolistsep]
  \item \textbf{Info-HCVAE}~\cite{lee-etal-2020-generating} leverages a hierarchical variational autoencoder to encode context paragraph. Latent variables for questions and answers are sampled from the latent distributions conditional on a context. During training, mutual information between question and answer representations are maximized to provide consistent QA pairs at test time.
  \item \textbf{AQGen}~\cite{shakeri-etal-2020-end} is a generative baseline model for question generation. We modify the AQGen architecture similar to the QAGen2S in the original paper, namely using encoder-decoders to separately generate answer and question based on the input context information.
  \item \textbf{QAGen} is based on QAGen2S in~\cite{shakeri-etal-2020-end}. Similar to AQGen, we exchange the task and generate possible questions end-to-end in the first step. Then, we generate answers based on both questions and contexts. 
\end{enumerate}

\subsection{Implementation}
\label{sec:implementation}
\textbf{Baselines:} For Info-HCVAE, AQGen, and QAGen, we apply roundtrip filtering and limited the maximum QA pairs per context to 5~\cite{alberti-etal-2019-synthetic}. Info-HCVAE is trained for $20$ epochs with the default settings~\cite{lee-etal-2020-generating}. For AQGen and QAGen, we implement the models based on~\cite{shakeri-etal-2020-end} and train for 10 epochs wit the learning rate of $1 \cdot 10^{-4}$ and the batch size of $16$. The optimizer is set to AdamW without weight decay and a linear warmup~\cite{loshchilov2017decoupled}, we validate the model with SQuAD dev set in training.

\textbf{QAGen-T5:} We apply LM-filtering as in~\cite{shakeri-etal-2020-end} and select QA pairs with highest scores for each context paragraph. QAGen-T5 models are trained similarly to AQGen and QAGen, we separately keep the best QG and QA models according to validation performance on the SQuAD dev set.

\textbf{QA model:} We follow~\cite{devlin-etal-2019-bert, kamath-etal-2020-selective} and train BERT-QA with learning rate of $3 \cdot 10^{-5}$ for two epochs and with a batch size of $16$. The AdamW optimizer is adopted and no linear warmup is used during training~\cite{loshchilov2017decoupled}.

\textbf{Hyperparameter search:} In our experiments, we empirically search for hyperparameters $\beta$ and $\sigma$ in the contrastive adaptation loss through additional experiments. We experiment with different values of $\beta$ in the range $[10^{-1}, 10^{-2}, 10^{-3}]$ and Gaussian noise $N(0, \sigma)$ applied on all token embeddings with standard deviation $\sigma$ ranging from $0$ to $10^{-2}$. The best combination of $\beta$ and $\sigma$ as per the training set is then selected, these numbers can be found in \Cref{tab:hyperparameter-selection}.

\begin{table}[ht]
  \footnotesize
  \centering
  \begin{tabular}{lcc}
  \hline
  \multirow{2}{*}{\textbf{Dataset}} & \textbf{Hyperparameter} & \textbf{CAQA Results} \\
                                    & $\beta$ / $\sigma$      & EM / F1               \\ \hline
  TriviaQA                          & 0.001/0.01              & 55.17/63.23           \\
  HotpotQA                          & 0.001/0.                & 46.37/61.57           \\
  NaturalQ.                         & 0.01/0.01               & 48.55/62.60           \\
  SearchQA                          & 0.001/0.01              & 36.05/42.94           \\ \hline
  \end{tabular}
  \caption[Hyperparameter selection]{Hyperparameter selection for each target dataset in our main results.}
  \label{tab:hyperparameter-selection}
\end{table}

All parameters that have not been mentioned explicitly above were used as reported in their original paper 

\section{Additional Results}
\label{section:additional-results}

\subsection{Comparison Limited Text Corpora vs. `Large' Text Corpora}

In this section, we compare our setting based on limited text corpora against the setting from the literature involving `large' text corpora. Hence, we report the results from (a)~the baseline models trained on SQuAD data (i.e., `SQuAD' as in our main paper), (b)~the baseline models using both SQuAD the 10k synthetic text corpora (i.e., `S + 10k Syn' as in our main paper) and (c) the baseline models using both SQuAD the all provided text corpora, results are from~\cite{lee-etal-2020-generating}. We also report (d), where $\sim$100k paragraphs are generated as synthetic QA data, which we take from~\cite{shakeri-etal-2020-end}. We refer to our implementation of (a) and (b) by marking the models using a `*'.

The results are in \Cref{tab:comparison-triviaqa} (TriviaQA) and \Cref{tab:comparison-naturalq} (NaturalQuestions). As expected, the setting (b) is responsible for a lower performance due to the limited text corpora. The performance in (b), as compared to (c) and (d), is lower by around $5\%$ to $10\%$. Importantly, our proposed \caqa still outperforms (b) by a considerable margin. Hence, despite using a considerable number sample of synthetic QA data, our \caqa is superior.           

\begin{table}[h!]
  \footnotesize
  \centering
  \begin{tabular}{lc}
  \hline
  \multirow{2}{*}{\textbf{Model}}            & \textbf{TriviaQA}          \\
                                             & EM / F1                    \\ \hline
  \multicolumn{2}{c}{Performance on Target Domain w/o Domain Adaptation}  \\ \hline
  (a) BERT-QA~\cite{lee-etal-2020-generating}    & 48.96/57.98                \\
  (a) BERT-QA*                                   & 50.84/60.48                \\ \hline
  \multicolumn{2}{c}{Performance on Target Domain w/ Question Generation} \\ \hline
  (c) HCVAE~\cite{lee-etal-2020-generating}      & 50.14/59.21                \\
  (b) HCVAE*                                     & 45.66/55.28                \\
  (b) QAGen*                                     & 50.52/59.79                \\ \hline
  \end{tabular}
  \caption[Comparison of results on TriviaQA]{BERT-QA and question generation results of our implementation and original work(s) on TriviaQA.}
  \label{tab:comparison-triviaqa}
\end{table}

\begin{table}[h!]
  \footnotesize
  \centering
  \begin{tabular}{lc}
  \hline
  \multirow{2}{*}{\textbf{Model}}           & \textbf{NaturalQ.}         \\
                                            & EM / F1                     \\ \hline
  \multicolumn{2}{c}{Performance on Target Domain w/o Domain Adaptation}  \\ \hline
  (a) BERT-QA~\cite{lee-etal-2020-generating}   & 42.77/57.29                 \\
  (a) BERT-QA~\cite{shakeri-etal-2020-end}      & 44.66/58.94                 \\
  (a) BERT-QA*                                  & 45.14/59.35                 \\ \hline
  \multicolumn{2}{c}{Performance on Target Domain w/ Question Generation} \\ \hline
  (c) HCVAE~\cite{lee-etal-2020-generating}     & 48.19/62.21                 \\
  (b) HCVAE*                                    & 37.12/51.17                 \\
  (d) QAGen~\cite{shakeri-etal-2020-end}        & 51.91/65.62                 \\
  (b) QAGen*                                    & 44.13/57.84                 \\ \hline
  \end{tabular}
  \caption[Comparison of results on NaturalQuestions]{BERT-QA and question generation results of our implementation and original work(s) on NaturalQuestions}
  \label{tab:comparison-naturalq}
\end{table}

\subsection{Sensitivity Analysis Varying the Number of QA Pairs per Context}
\label{sec:varying_qa_per_context}
We now perform a sensitivity analysis in which we vary the number of QA pairs per context (i.e., $k$). For this, we again adopt our \caqa framework (with both QAGen-T5 model and contrastive adaptation loss) using a combination of the SQuAD dataset and 10k context paragraphs.  We vary the number of QA pairs for each context in the range $k = 1$, 3, 5, 7, and 9 QA pairs. The results are presented in \Cref{tab:qa-per-context-number-results}. We note only some minor variation. The improvements tend to be larger when increasing the number of QA pairs per context in HotpotQA, while the results for SearchQA are less stable when increasing the number of of synthetic QA data.

\begin{table}[ht]
  \footnotesize
  \centering
  \begin{tabular}{lcc}
  \hline
  \multirow{2}{*}{\textbf{Training data}} & \textbf{HotpotQA}    & \textbf{SearchQA}    \\
                                          & EM / F1              & EM / F1              \\ \hline
  \multicolumn{3}{c}{Performance w/o domain adaptation}                                 \\ \hline
  SQuAD                                   & 43.57/61.09          & 19.66/27.90          \\ \hline
  \multicolumn{3}{c}{Performance w/ question generation}                                \\ \hline
  S + 1QA/C                               & 45.03/60.39          &  32.40/40.29          \\
  S + 3QA/C                               & 45.92/61.13          & 34.96/42.47          \\
  S + 5QA/C                               & 46.50/62.03          & 32.54/39.44          \\
  S + 7QA/C                               & 47.45/62.83          & \textbf{37.05/44.46} \\
  S + 9QA/C                               & \textbf{47.57/63.26} & 33.10/40.12          \\
  S + 5QA/C /w CAQA                       & 46.37/61.57          & 36.05/42.94          \\ \hline
  \end{tabular}
  \caption[Sensitivity analysis varying the number of QA pairs per context]{Sensitivity analysis varying the number of QA pairs per context ($k$).}
  \label{tab:qa-per-context-number-results}
\end{table}

\newpage
\FloatBarrier
\section{Qualitative Analysis of Synthetic Data Samples}
\label{sec:qualitative-examples}

We present qualitative examples of generated synthetic QA data using the proposed \caqa framework with our QAGen-T5 model. For this, two context paragraphs and five QA pairs for each paragraph are presented in the following; see \Crefrange{tab:triviaqa-examples}{tab:searchqa-examples}.

\begin{table*}[h]
  \caption[Synthetic TriviaQA samples]{Synthetic TriviaQA samples generated by QAGen-T5.}
  \centering
  \begin{tabular}{p{0.95\linewidth}}
  \hline
  \multicolumn{1}{c}{Synthetic TriviaQA samples} \\ \hline
  \textbf{Context}: \small{UEFA European Championship Top 10 Midfielders - Football BibleUEFA European Championship Top 10 Midfielders  UEFA European Championship Top 10 Midfielders  Published on  Country: Portugal  Years Active: 1990 to 2008  Rui Costa represented Portugal in three UEFA European Championships . He was a good provider for other players to score goals but still scored a good number himself. At his time, the national side had a steady performance making it to the quarter finals in 1996, then the semifinals at the Euro 2000 and eventually the Euro 2004 final. On the progress to the final, he helped the Portugal team to eliminate England at the quarter finals through a remarkable strike which secured a 2-1 win......} \\
  \textbf{QA 1}: What country did Rui Costa represent?    Portugal \\
  \textbf{QA 2}: In what year did Iniesta win UEFA Team of the Tournament?    1988 \\
  \textbf{QA 3}: What team did he play for in the European Football Championships?    Portugal \\
  \textbf{QA 4}: Who is the second highest assist-getter in La Liga?    Luis Figo \\
  \textbf{QA 5}: Who was a good provider for other players to score goals?    Rui Costa \\ \hline
  \textbf{Context}: \small{Spiers on Sport: the unjust sacking of Kenny Shiels (From ...Spiers on Sport: the unjust sacking of Kenny Shiels (From HeraldScotland)  / Spiers on Sport , Graham Spiers  When a manager wins one of only four trophies collected by a football club in 80 years, there has to be a degree of respect shown towards him, right?  When he also works slavishly on all aspects of a club due to staffing limitations - training, recruiting, video-editing, youth development etc - wouldn't that admiration for him grow even greater?  Loading article content  Kenny Shiels, sacked by Kilmarnock, is by no means perfect. But he has been a pretty good manager at Rugby Park, whose dismissal is hard to fathom......} \\
  \textbf{QA 1}: Where did most of Shiels' felonies occur? Rugby Park \\
  \textbf{QA 2}: What club did he manage?    Kilmarnock \\
  \textbf{QA 3}: Who is the chairman of the rugby club?    Michael Johnston \\
  \textbf{QA 4}: What was the name of the team that he managed?    Kilmarnock \\
  \textbf{QA 5}: Who was sacked by Kilmarnock?    Kenny Shiels \\ \hline
  \end{tabular}
  \label{tab:triviaqa-examples}
\end{table*}

\begin{table*}[h]
  \caption[Synthetic HotpotQA samples]{Synthetic HotpotQA samples generated by \caqa.}
  \centering
  \begin{tabular}{p{0.95\linewidth}}
  \hline
  \multicolumn{1}{c}{Synthetic HotpotQA samples} \\ \hline
  \textbf{Context}: \small{Cascade Range {[}SEP{]} The Cascade Range or Cascades is a major mountain range of western North America, extending from southern British Columbia through Washington and Oregon to Northern California.  It includes both non-volcanic mountains, such as the North Cascades, and the notable volcanoes known as the High Cascades.  The small part of the range in British Columbia is referred to as the Canadian Cascades or, locally, as the Cascade Mountains.  The latter term is also sometimes used by Washington residents to refer to the Washington section of the Cascades in addition to North Cascades, the more usual U.S. term, as in North Cascades National Park.  The highest peak in the range is Mount Rainier in Washington at 14411 ft......} \\
  \textbf{QA 1}: What is one of Oregon's most popular outdoor recreation sites?    Lake of the Woods \\
  \textbf{QA 2}: Who named the island?    Oliver C. Applegate \\
  \textbf{QA 3}: What is the name of the lake in Oregon?    Lake of the Woods \\
  \textbf{QA 4}: What is the name of the unincorporated community located on the east shore of the lake? Lake of the Woods \\
  \textbf{QA 5}: What is another name for the Cascade Range?    Cascades \\ \hline
  \textbf{Context}: \small{Jim Conroy {[}SEP{]} James Conroy (born February 6, 1977) is an American voice actor, television writer and actor.  He is known for appearing on television shows, such as "Celebrity Deathmatch", "Kenny the Shark" and "Fetch with Ruff Ruffman", radio commercials and video games.  He worked for companies such as WGBH, The Walt Disney Company and Discovery Channel. {[}PAR{]} {[}TLE{]} Kenny the Shark {[}SEP{]} Kenny the Shark is an American animated television series produced by Discovery Kids. The show premiered on NBC's Discovery Kids on NBC from November 1, 2003 and ended February 18, 2006 with two seasons and 26 episodes in total having aired......} \\
  \textbf{QA 1}: How many episodes did the show have?    26 \\
  \textbf{QA 2}: What is the birth date of Jim Conroy?    February 6, 1977 \\
  \textbf{QA 3}: What is the name of the American animated television series?    Kenny the Shark \\
  \textbf{QA 4}: Who produces Kenny the Shark?    Discovery Kids \\
  \textbf{QA 5}: What is Jim Conroy's birth date?    February 6, 1977 \\ \hline
  \end{tabular}
  \label{tab:hotpotqa-examples}
\end{table*}

\begin{table*}[h]
  \caption[Synthetic NaturalQuestions samples]{Synthetic NaturalQuestions samples generated by \caqa.}
  \centering
  \begin{tabular}{p{0.95\linewidth}}
  \hline
  \multicolumn{1}{c}{Synthetic NaturalQuestion samples} \\ \hline
  \textbf{Context}: \small{\textless{}Table\textgreater \textless{}Tr\textgreater \textless{}Th\textgreater Rank \textless{}/Th\textgreater \textless{}Th\textgreater Chg \textless{}/Th\textgreater \textless{}Th\textgreater Channel name \textless{}/Th\textgreater \textless{}Th\textgreater Network \textless{}/Th\textgreater \textless{}Th\textgreater Primary language ( s ) \textless{}/Th\textgreater \textless{}Th\textgreater Subscribers ( millions ) \textless{}/Th\textgreater \textless{}Th\textgreater Content category \textless{}/Th\textgreater \textless{}/Tr\textgreater \textless{}Tr\textgreater \textless{}Td\textgreater 1 . \textless{}/Td\textgreater \textless{}Td\textgreater \textless{}/Td\textgreater \textless{}Td\textgreater PewDiePie \textless{}/Td\textgreater \textless{}Td\textgreater N / A \textless{}/Td\textgreater \textless{}Td\textgreater English \textless{}/Td\textgreater \textless{}Td\textgreater 64 \textless{}/Td\textgreater \textless{}Td\textgreater Comedy \textless{}/Td\textgreater \textless{}/Tr\textgreater \textless{}Tr\textgreater \textless{}Td\textgreater 2 . \textless{}/Td\textgreater \textless{}Td\textgreater \textless{}/Td\textgreater \textless{}Td\textgreater T - Series \textless{}/Td\textgreater \textless{}Td\textgreater T - Series \textless{}/Td\textgreater \textless{}Td\textgreater Hindi \textless{}/Td\textgreater \textless{}Td\textgreater 53 \textless{}/Td\textgreater \textless{}Td\textgreater Music \textless{}/Td\textgreater \textless{}/Tr\textgreater \textless{}Tr\textgreater \textless{}Td\textgreater 3 . \textless{}/Td\textgreater \textless{}Td\textgreater \textless{}/Td\textgreater \textless{}Td\textgreater Justin Bieber \textless{}/Td\textgreater \textless{}Td\textgreater N / A \textless{}/Td\textgreater \textless{}Td\textgreater English \textless{}/Td\textgreater \textless{}Td\textgreater 40 \textless{}/Td\textgreater \textless{}Td\textgreater Entertainment \textless{}/Td\textgreater \textless{}/Tr\textgreater \textless{}Tr\textgreater \textless{}Td\textgreater 4 . \textless{}/Td\textgreater \textless{}Td\textgreater \textless{}/Td\textgreater \textless{}Td\textgreater Canal KondZilla......} \\
  \textbf{QA 1}: What language is used in the Disney films?    Hindi \\
  \textbf{QA 2}: What company is Sony Pictures?    Sony Pictures \\
  \textbf{QA 3}: What is the name of the band that released the single "One Direction"?    One Direction \\
  \textbf{QA 4}: What is the name of the band that is known for its trap music?    PewDiePie \\
  \textbf{QA 5}: What is Katy Perry's real name?    Katy Perry \\ \hline
  \textbf{Context}: \small{Red blood cell distribution width ( RDW or RDW - CV or RCDW and RDW - SD ) is a measure of the range of variation of red blood cell ( RBC ) volume that is reported as part of a standard complete blood count . Usually red blood cells are a standard size of about 6 - 8 \textbackslash{}u03bcm in diameter . Certain disorders , however , cause a significant variation in cell size . Higher RDW values indicate greater variation in size . Normal reference range of RDW - CV in human red blood cells is 11.5 - 14.5 \% . If anemia is observed , RDW test results are often used together with mean corpuscular volume ( MCV ) results to determine the possible causes of the anemia . It is mainly used to differentiate an anemia of mixed causes from an anemia of a single cause......} \\
  \textbf{QA 1}: What do higher RDW values indicate?    Greater variation in size \\
  \textbf{QA 2}: What is the measure of the range of variation of red blood cell volume?    Red blood cell distribution width \\
  \textbf{QA 3}: What can cause a significant variation in cell size?    Certain disorders \\
  \textbf{QA 4}: What is the normal reference range of RDW - CV in human red blood cells?    11.5 - 14.5 \% \\
  \textbf{QA 5}: What is the main purpose of the test?    To differentiate an anemia of mixed causes from an anemia of a \\ \hline
  \end{tabular}
  \label{tab:naturalq-examples}
\end{table*}

\begin{table*}[h]
  \caption[Synthetic SearchQA samples]{Synthetic SearchQA samples generated by \caqa.}
  \centering
  \begin{tabular}{p{0.95\linewidth}}
  \hline
  \multicolumn{1}{c}{Synthetic SearchQA samples} \\ \hline
  \textbf{Context}: \small{A white elephant - Idioms by The Free Dictionary {[}PAR{]} Definition of a white elephant in the Idioms Dictionary. a white elephant phrase. What does a white elephant expression mean? Definitions by the largest Idiom... {[}DOC{]} {[}TLE{]} Can an elephant stand up after laying down? {[}Archive{]} - Straight ... {[}PAR{]} We often receive e-mails from avid EleCam viewer saying, "There are elephants lying down in the pasture. They have been there a long time. {[}DOC{]} {[}TLE{]} Elephants sleep in zoo and circus {[}PAR{]} That is one of the reasons why elephants do not sleep much, and then only with ... The first elephant starts to lie down on its side towards 11 o'clock at night. {[}DOC{]} {[}TLE{]} Elephant Who Gives Rides All Day Can't Even Lie Down To Rest......} \\
  \textbf{QA 1}: What can an elephant do after lying down?    Stand up \\
  \textbf{QA 2}: What do I struggle with?    People who lie \\
  \textbf{QA 3}: What is the name of the elephant who gives rides all day?    Elephant Who Gives Rides All Day Can't Even Lie Down \\
  \textbf{QA 4}: What is the official website of South African National Parks?    SANParks \\
  \textbf{QA 5}: What is more concerning to me than lies?    Misbehavior \\ \hline
  \textbf{Context}: \small{jeopardy/1333\_Qs.txt at master  jedoublen/jeopardy  GitHub {[}PAR{]} Number: 2. ANIMAL SONGS | British singer Robyn Hitchcock is known for his tunes about these animals, including "Bass" \& "Aquarium" | Fish. right: Matt. Wrong:. {[}DOC{]} {[}TLE{]} Robyn Hitchcock - Wikipedia {[}PAR{]} Robyn Rowan Hitchcock (born 3 March 1953) is an English singer-songwriter and guitarist. While primarily a vocalist and guitarist, he also plays harmonica, piano, and bass guitar. ... Hitchcock's lyrics tend to include surrealism, comedic elements, ... Hitchcock released his solo debut, Black Snake Diamond Rle in 1981,... {[}DOC{]} {[}TLE{]} Positive Vibrations: Softcore - fegMANIA! {[}PAR{]} An except from Positive Vibrations' complete guide to the songs of Robyn Hitchcock......} \\
  \textbf{QA 1}: What is the dance music of northeastern Argentina known as?    Chaman \\
  \textbf{QA 2}: What was Hitchcock's solo debut called?    Black Snake Diamond Rle \\
  \textbf{QA 3}: When did Hitchcock release his solo debut?    1981 \\
  \textbf{QA 4}: What is the name of the book that contains a complete guide to the songs of Robyn Hitchcock?    Positive Vibrations: Softcore - fegMA \\
  \textbf{QA 5}: What was the name of the singer who performed on The House List?    Robyn Hitchcock \\ \hline
  \end{tabular}
  \label{tab:searchqa-examples}
\end{table*}

\newpage
\FloatBarrier
\section{PCA Visualization of Data}
\label{sec:pca-examples}
\begin{figure*}[h]
  \centering
  \includegraphics[width=0.95\textwidth]{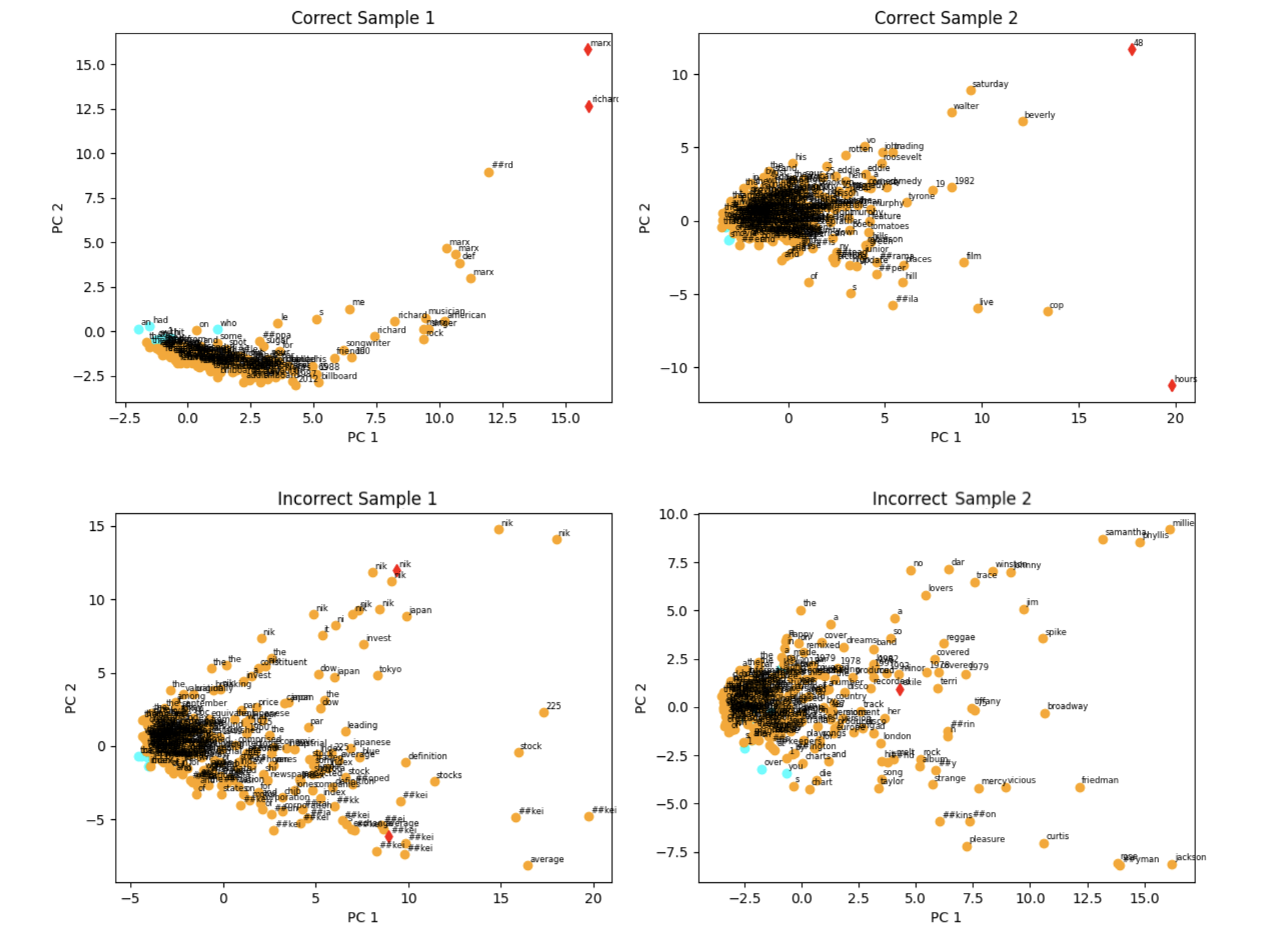}
  \caption{Visualization of BERT-encoder output on TriviaQA w/ contrastive adaptation loss.}
  \label{fig:triviaqa-contrastive}
\end{figure*}

\begin{figure*}[h]
  \centering
  \includegraphics[width=0.95\textwidth]{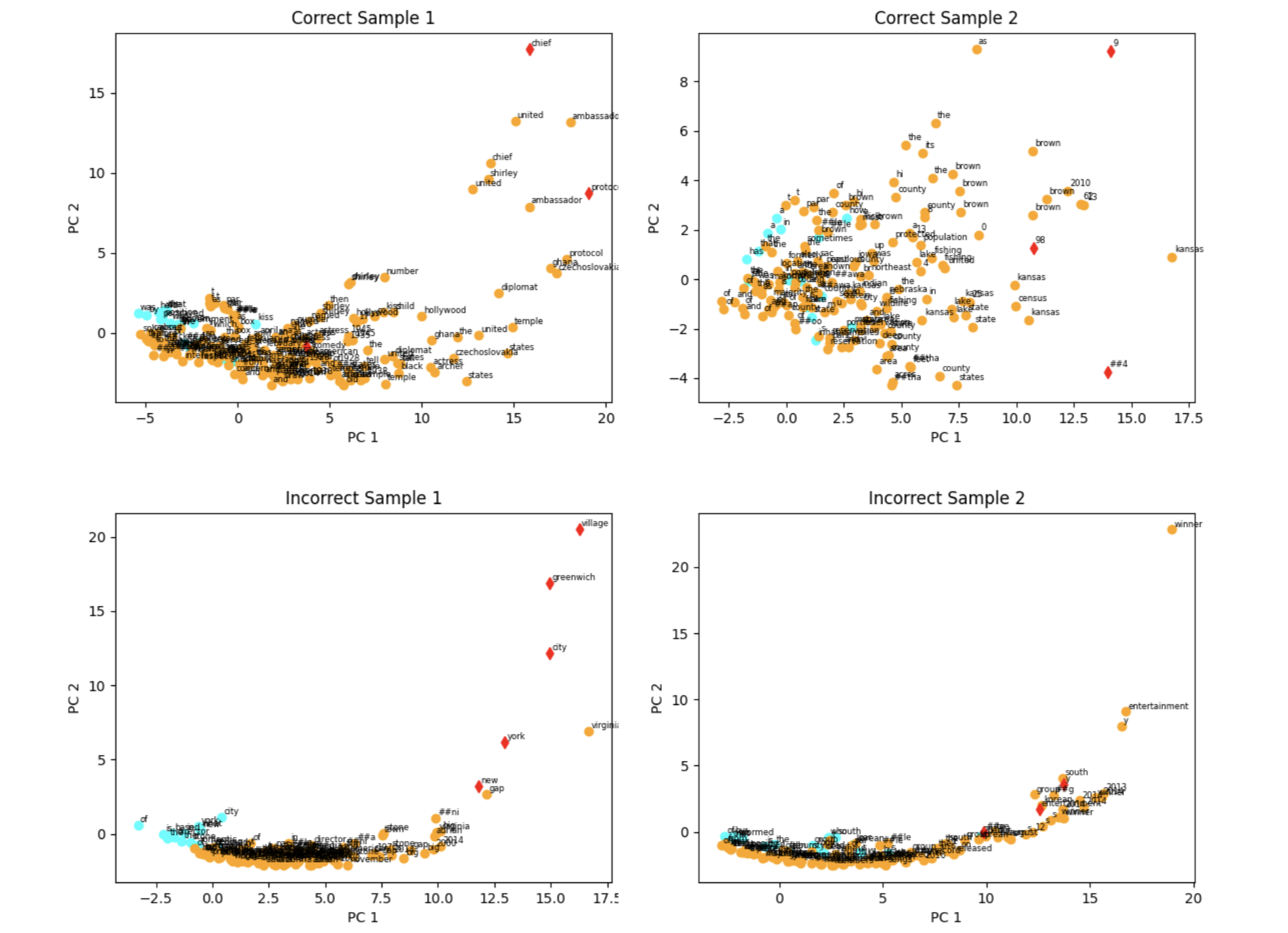}
  \caption{PCA visualization of BERT-encoder output on HotpotQA w/ contrastive adaptation loss.}
  \label{fig:hotpotqa-contrastive}
\end{figure*}

\begin{figure*}[h]
  \centering
  \includegraphics[width=0.95\textwidth]{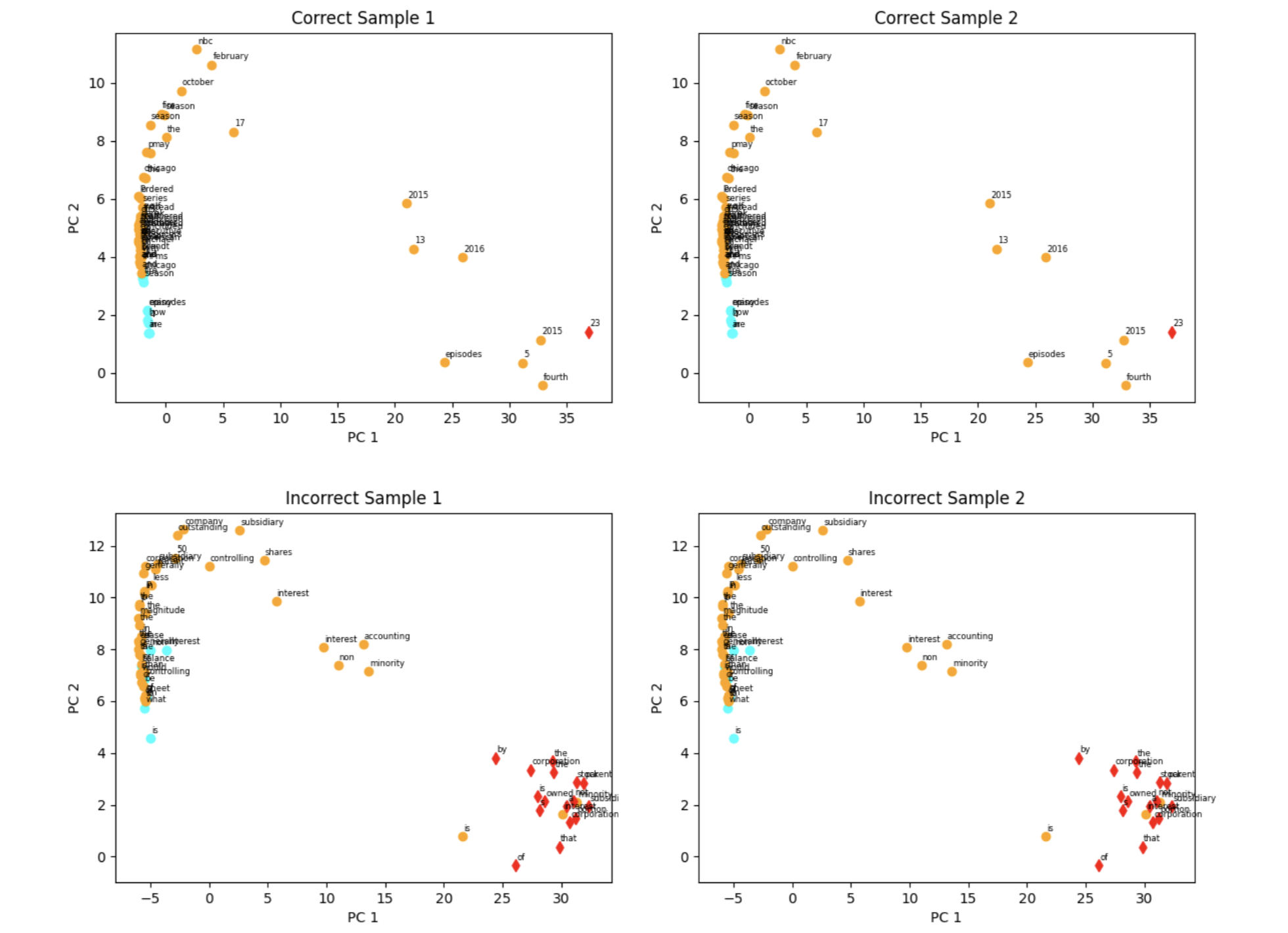}
  \caption{PCA visualization of BERT-encoder output on NaturalQuestions w/ contrastive adaptation loss.}
  \label{fig:naturalq-contrastive}
\end{figure*}

\begin{figure*}[h]
  \centering
  \includegraphics[width=0.95\textwidth]{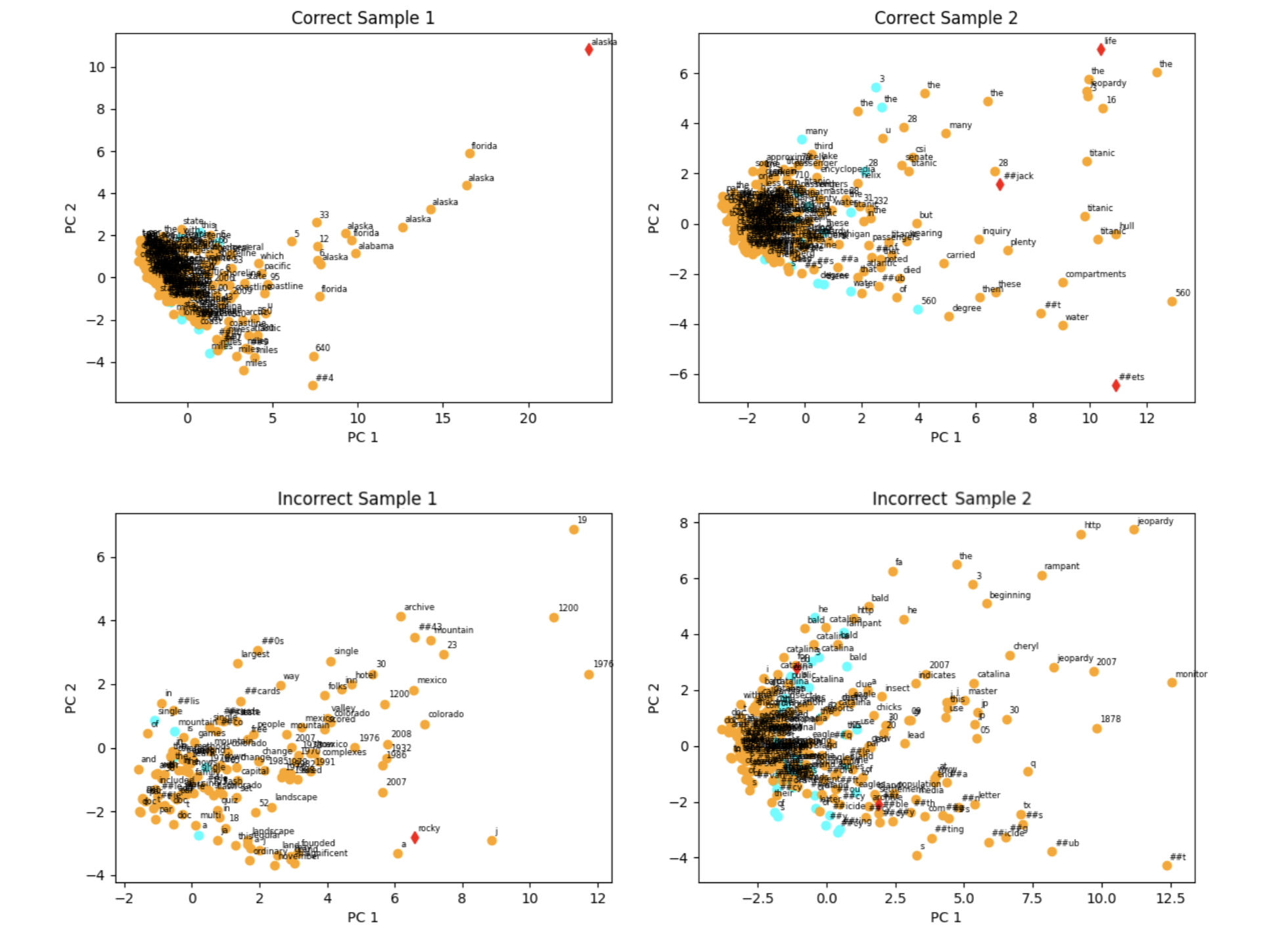}
  \caption{PCA visualization of BERT-encoder output on SearchQA w/ contrastive adaptation loss.}
  \label{fig:searchqa-contrastive}
\end{figure*}

We visualize the BERT-QA output for the synthetic QA data generated by our QAGen-T5 model. Here, BERT-QA models are trained with contrastive adaptation loss on all target datasets separately. The results are shown for TriviaQA (\Cref{fig:triviaqa-contrastive}), HotpotQA (\Cref{fig:hotpotqa-contrastive}), NaturalQuestions (\Cref{fig:naturalq-contrastive}), and SearchQA (\Cref{fig:searchqa-contrastive}). Answer tokens are in red diamond shapes, question tokens in cyan circles, while all other tokens are represented in orange circles.

\end{document}